\documentclass[letterpaper, 10 pt, journal, twoside]{IEEEtran}
\IEEEoverridecommandlockouts                              

\usepackage[ruled, lined, linesnumbered, commentsnumbered, longend]{algorithm2e}
\usepackage{array,multirow}
\usepackage{graphicx}
\usepackage[caption=false,font=small,labelfont=sf,textfont=sf]{subfig}
\usepackage[font=small,labelfont=bf]{caption}
\usepackage{textcomp}
\usepackage{stfloats}
\usepackage{url}
\usepackage{verbatim}
\usepackage{cite}
\usepackage{tabularx}
\usepackage{xcolor}
\usepackage{siunitx}
\usepackage{makecell}
\usepackage{comment}
\usepackage{float}
\usepackage{booktabs}
\usepackage{algpseudocode}
\usepackage{mathtools}
\usepackage{bm}
\usepackage{diagbox}
\usepackage{epsfig} 
\usepackage{times}
\usepackage{listings}
\usepackage{psfrag}
\usepackage{dirtytalk}
\usepackage[colorlinks,linkcolor=black,citecolor=black,urlcolor=black,bookmarks=false,hypertexnames=true]{hyperref} 
\usepackage{xparse}
\usepackage[english]{babel}
\usepackage{multirow}
\usepackage{amsmath,amsfonts}
\usepackage{amssymb}
\newcommand{\emphb}[1]{\textbf{\emph{#1}}}

\title{\LARGE \bf
Robust Statistics vs. Machine Learning vs. Bayesian Inference: \\ 
Insights into Handling Faulty GNSS Measurements \\ for Robust Vehicle Localization
}

\author{Haoming Zhang,~\IEEEmembership{Member,~IEEE}
\thanks{Haoming Zhang was with the Institute of Automatic Control, RWTH Aachen University, Aachen, Germany, and now with the Learning Systems and Robotics Lab, Technical University of Munich, Munich, Germany, and with the Munich Institute of Robotics and Machine Intelligence (MIRMI).}%
\thanks{Corresponding: haoming.zhang@tum.de}%
}

\usepackage{siunitx}
\usepackage{graphicx}
\usepackage{multirow}
\usepackage{comment}
\usepackage{bm}
\usepackage{hhline}
\usepackage[font=small,labelfont=bf]{caption}
\usepackage{macros}
\DeclareMathAlphabet\mathbfcal{OMS}{cmsy}{b}{n}

\begin{document}

\IEEEaftertitletext{\vspace{-1.\baselineskip}}
\maketitle
\thispagestyle{empty}
\pagestyle{empty}

\begin{abstract}
This paper presents research findings on handling faulty measurements (i.e., outliers) of global navigation satellite systems (GNSS) for vehicle localization under adverse signal conditions in field applications, where raw GNSS data are frequently corrupted due to environmental interference such as multipath, signal blockage, or non-line-of-sight conditions. In this context, we investigate three strategies applied specifically to GNSS pseudorange observations: robust statistics for error mitigation, machine learning for faulty measurement prediction, and Bayesian inference for noise distribution approximation. 

Since previous studies have provided limited insight into the theoretical foundations and practical evaluations of these three methodologies within a unified problem statement (i.e., state estimation using ranging sensors), we conduct extensive experiments using real-world sensor data collected in diverse urban environments. 

Our goal is to examine both established techniques and newly proposed methods, thereby advancing the understanding of how to handle faulty range measurements, such as GNSS, for robust, long-term vehicle localization. In addition to presenting successful results, this work highlights critical observations and open questions to motivate future research in robust state estimation.
\end{abstract}

\section{Motivation}
Vehicle navigation in \emphb{large-scale field environments} commonly depends on Global Navigation Satellite Systems (GNSS) to maintain globally consistent state estimates. However, in complex environments such as urban canyons and forest, GNSS signals are frequently degraded by multipath propagation and non-line-of-sight (NLOS) conditions. These effects lead to measurement outliers with time-varying noise characteristics that are often \emphb{heavy-tailed}, \emphb{skewed}, and \emphb{multimodal}, making them difficult to model a priori \cite{hsu_analysis_NLOS}. 

Consequently, state estimators based on least-squares techniques may degrade or even diverge when relying on inconsistent measurement models with overconfident, hand-crafted noise parameters, as they oftentimes underestimate the complexity and magnitude of uncertainty. As a result, developing a vehicle localization approach that ensures robust \emphb{long-term operation} across diverse environments remains a significant challenge. 

\subsection{Related Work}
Given that faulty sensor observations pose a critical challenge to trustworthy autonomy in field applications, state-of-the-art methods typically address this issue by leveraging five key concepts:

\subsubsection{\textbf{Multisensor Fusion}}\label{sec: literature_fusion}
Many approaches emphasize fusing additional sensor modalities alongside GNSS observations to enhance state observability in the presence of signal loss or to mitigate the effects of faulty measurements. Previous studies have extensively explored the integration of various local sensors into graph-based optimization frameworks, such as near-field range signals \cite{gnss_uwb}, lidar \cite{liosam}, cameras \cite{gvins, wen_vis_gnss}, or multiple sensors \cite{superodom}, alongside GNSS for state estimation, particularly in SLAM applications. 

In our previous work \cite{fgo_zhang, gnssfgo}, we proposed a generalized GNSS/multisensor state estimator based on continuous-time factor graph optimization by incorporating Gaussian process motion priors, originally introduced in \cite{WNOJ}. The flexibility comes from the fact that we choose estimation timestamps independent of any particular sensor frequency. This feature, as well as the smoothing effect of a motion prior, provides robustness in the presence of any particular sensor dropout. 

\textbf{Limitations:}
Even when redundant state constraints are available through multiple sensor inputs, the state estimators reviewed above still assume that measurement noise follows a Gaussian distribution. This assumption tends to underestimate the complexity and magnitude of real-world uncertainty. As a result, they often struggle to effectively downweight faulty sensor observations. Without advanced engineering and fine-tuned parameters, the performance of multisensor fusion systems is compromised, limiting their scalability and reliability for long-term operations in large-scale environments.

Therefore, robust state estimation for handling faulty measurements (outliers) remains an active area of research.

\subsubsection{\textbf{Outlier Exclusion}}\label{sec:literature_outlier}
Outlier exclusion serves as a discriminative strategy for handling measurement outliers in state estimation problems. Rather than relying on statistical modeling of the outlier distribution, this approach uses heuristic methods to detect and exclude faulty GNSS measurements. 

Methods in this category typically employ consensus checking (e.g., RANSAC) \cite{ransac_gnss_Zhang} and integrity monitoring techniques \cite{IM_GNSS_review}, where two further principles are commonly employed: statistical testing \cite{raim_multiple}, which evaluates measurement consistency under redundancy, and interval analysis \cite{gnss_interval_Drevelle, gnss_zonotope}, which bounds uncertainties through conservative set representations to detect anomalies.

\textbf{Limitations:}
RANSAC-like algorithms are known to become inefficient and unstable when dealing with high outlier rates ($> \SI{50}{\%}$) \cite{ransac_downside}, and they struggle to scale effectively in high-dimensional problems \cite{outlier_spatial_perception}. Similarly, integrity monitoring assumes sufficient healthy measurements to maintain consistency. When outliers dominate or too few measurements are available, the estimated protection level becomes unreliable.

\subsubsection{\textbf{Robust Error Modeling}}\label{sec: literature_error_modeling}
Robust error modeling presents another effective approach to outlier rejection, commonly achieved by shaping the cost function to enhance resilience against outliers.

In this category, M-estimation (maximum likelihood type estimates) has received considerable attention for rejecting GNSS measurement outliers \cite{Comparison_Robust_Estimation_GPS_Knight, app_gnss_m_est_Crespillo}. Their utility has also been validated in harsh environments with significant multipath and NLOS effects \cite{medina_on_robust_statistics, Chauchat2024}. 

While M-estimation is well-established, its effectiveness relies on selecting a well-parameterized kernel function. Recent work has focused on adaptive M-estimation, where parameters are tuned based on current error statistics. \cite{M-est_self_learning_Ding} proposed a self-learning Huber loss using GNSS residuals, showing slight performance gains. \cite{DasScale-Variant} introduced a two-stage method to tune the generalized kernel from \cite{M-est_generalization_Barron}, leveraging past measurements to adapt to outlier behavior.

\textbf{Limitations:} Most studies using non-adaptive M-estimators focus on hard-redescending kernels like Huber and Tukey, overlooking the potential of soft-redescending functions for robust GNSS-based vehicle localization. They also lack analysis of parameter tuning and the trade-off between robustness and efficiency. Although some works \cite{ZHANG199759, de2021review, M-estimate_eva_MacTavish} evaluate various kernels and settings, they are not tailored to range-based localization.

Adaptive M-estimation methods are typically limited to simple, symmetric noise models and rely on past residuals, which may fail when noise patterns shift abruptly. In low-visibility GNSS scenarios, estimator efficiency becomes as crucial as robustness.

These gaps highlight the need for a comprehensive study of M-estimators across categories, focusing on their mathematical properties and suitability for robust GNSS-based vehicle localization.



\subsubsection{\textbf{Bayesian Inference for Distribution Approximation}}
Another recent work has tackled measurement outliers by integrating robust error modeling into the Variational Bayes framework. In simpler cases, measurement noise is modeled with a conjugate prior (e.g., Gaussian) from the exponential family, enabling efficient inference \cite{SE_VBI_Courts, GVBI_Goudar}.

When noise exhibits complex characteristics, such as heavy tails, multimodality, or asymmetry, beyond the capacity of exponential-family priors, mixture models serve as universal approximators. Gaussian mixture models, in particular, have been widely applied to fit noise distributions from measurement residuals, effectively capturing diverse noise patterns \cite{phd_Pfeifer}.

\textbf{Limitations:}
However, approximated noise distributions are often complex, nonlinear, and non-convex, making it difficult to extract reliable covariance estimates for least-squares solvers, potentially leading to suboptimal results.

Moreover, sampling measurement residuals poses challenges due to the lack of ground-truth errors. While prior work \cite{Wenda_VBIGMM} addresses state uncertainty in residuals, distribution shifts can still occur, especially in dynamic, online settings like vehicle motion. Additional challenges, such as sampling efficiency and its context awareness, remain open research questions.

\subsubsection{\textbf{Learning-based Methods}}\label{sec: literature_learning}
As environment-dependent outliers may be challenging to identify and can exhibit complex noise patterns that are difficult to model with traditional statistical methods, learning-based methods\footnote{Although many learning-based methods are rooted in statistical inference and thus share similar theoretical foundations, we intentionally categorize inference-based and learning-based methods as distinct approaches. 
} have also gained attention for handling faulty GNSS observations. 

Pre-trained models enable online GNSS fault detection. While classic machine-learning methods can also achieve high accuracy \cite{classical_learning_1, classical_learning_2}, deep learning (DL) better handles time-correlated, nonlinear GNSS data by extracting semantic features and improving generalization \cite{learning_survey, lstm_base}. In our previous work \cite{te_lstm}, we proposed a novel neural network architecture to predict measurement outliers following the noise characteristics of the GNSS observations. In addition, we provided an in-depth analysis comparing classical models with deep networks, as well as evaluating performance on in-distribution and out-of-distribution data.

\textbf{Limitations:} However, while previous studies highlight the potential of learning-based methods for handling GNSS outliers, they have not thoroughly evaluated their real-world feasibility, limitations, or generalization across diverse datasets and environments. 

Moreover, as there is currently no evidence that learning-based methods can generalize across varying scenarios in vehicle localization, further investigation is needed into how learning concepts can be effectively integrated to enhance the robustness of state estimation applications.

\subsection{Objective}
Recognizing the limitations of the aforementioned studies, this paper summarizes the author's key research findings and lessons learned on handling faulty GNSS measurements and aims to evaluate the feasibility and limitations of three prominent method categories using data from real-world measurement campaigns.

In particular, we outline the following objectives:
\begin{itemize}
    \item \textbf{Analyze various M-estimators} for GNSS-based localization, aiming to develop a systematic understanding of how different kernel functions balance robustness and efficiency.
    
    \item \textbf{Evaluate offline learning-based methods}, including both classical machine learning models and deep neural networks, to predict faulty GNSS measurements. We provide an in-depth comparison between these model types and assess their performance on both in-distribution and out-of-distribution datasets.

    \item \textbf{Validate online variational Bayesian inference} for approximating GNSS pseudorange noise distributions, with a particular focus on distribution shifts in residual samples and identifying remaining open challenges.
\end{itemize}

\section{Preliminaries}
\subsection{GNSS Pseudorange Model for Vehicle Localization}
In localization approaches that tightly fuse the GNSS observations, pseudorange is commonly used \cite{groves}. The pseudorange $\myFrameScalar{\rho}{}{}$ represents a geometric distance between the phase center of the GNSS antenna and the associated satellite, which contains several range delays due to satellite orbit bias and atmospheric delays. It can be modeled with respect to the antenna position as
\begin{equation}
    \myFrameScalar{\rho}{k}{}= \underbrace{\left\|\myFrameVec{p}{\rm ant}{e}-\myFrameVec{p}{\mathrm{sat},k}{e}\right\|}_{\text{1-D geometric range}}
                        + \underbrace{\myFrameScalar{c}{b}{} - c_{b, \mathrm{sat}} + T + I}_{\text{assumed to be known}} + \underbrace{M + w_{\rho,k}}_{\text{\textcolor{red}{unknown (Fig.1a)}}}\mathrm{,}
\label{eq: rho}
\end{equation}
where the vectors $\myFrameVec{p}{\rm ant}{e}$ and $\myFrameVec{p}{\mathrm{sat},k}{e}$ represent the positions of the GNSS antenna and $k$-th satellite in the Earth-centered, Earth-fixed (ECEF) frame, respectively. The variables $\myFrameScalar{c}{b}{}$ and $c_{b, \mathrm{sat}}$ represent the bias due to receiver clock delay and satellite clock delay. The tropospheric, ionospheric and multipath delays are denoted as $T(t)$, $I(t)$, and $M$. The pseudorange noise is $w_{\rho}$.

Throughout this research, we assume that all clock errors and atmospheric delays can be estimated or eliminated using additional correction data.

\subsection{GNSS Measurement Corruption}
\begin{figure}[!h]
    \centering
    \includegraphics[width=0.48\textwidth]{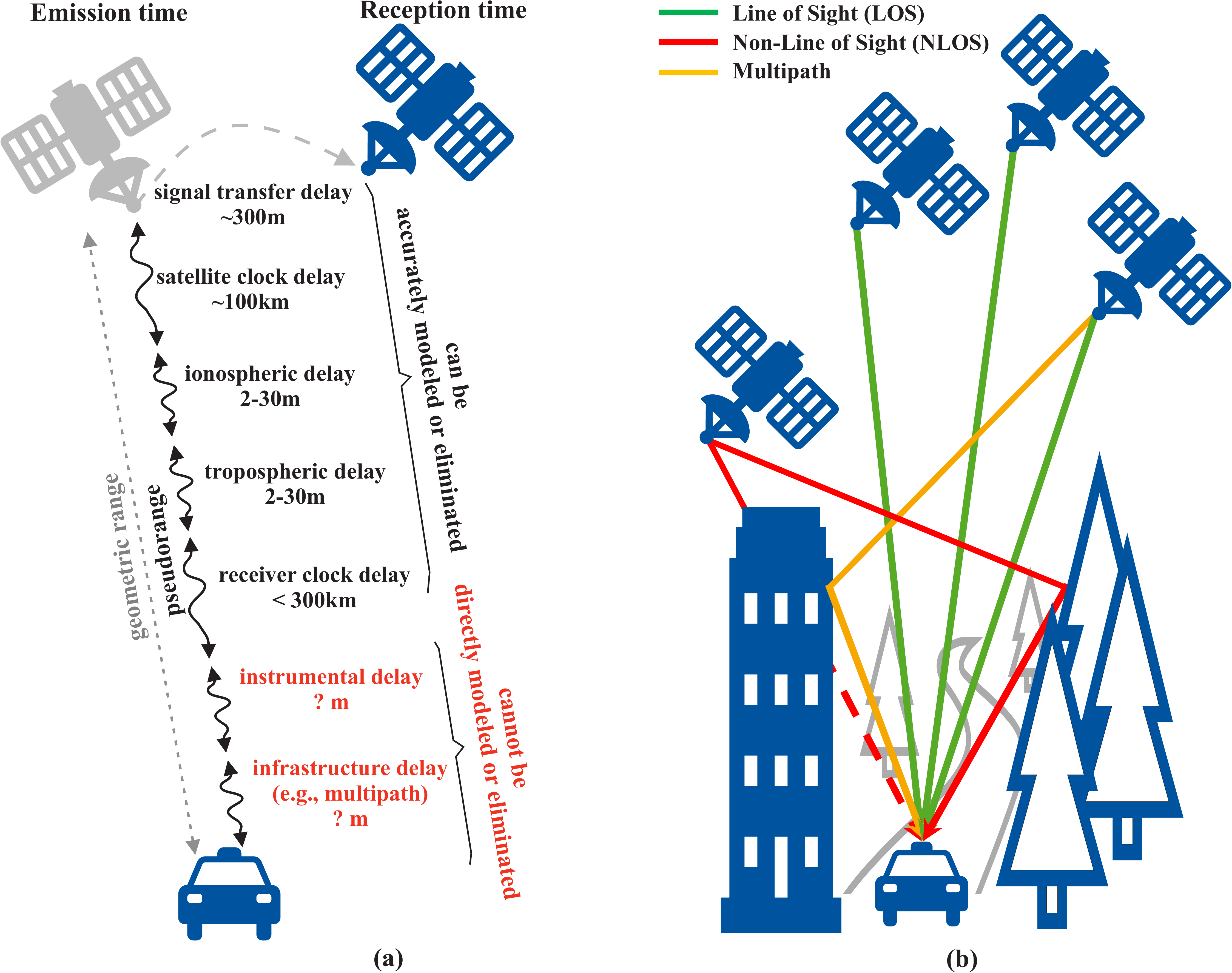}
    \caption{GNSS delays and interference. }
    \label{fig: gnss_delays}
\end{figure}

Fig.\,\ref{fig: gnss_delays} presents an overview of GNSS measurement corruption, illustrating various sources of error in pseudorange observations. These include delays induced by satellite and receiver clocks, signal propagation through the ionosphere and troposphere, as well as instrumental and infrastructure-related effects (see Fig.\,\ref{fig: gnss_delays} a). While some of these errors, such as clock offsets or atmospheric delays, can be accurately eliminated, others, like multipath and NLOS receptions, introduce non-deterministic disturbances that are difficult to eliminate. Fig.\,\ref{fig: gnss_delays}b further distinguishes between line-of-sight (LOS) and NLOS scenarios, highlighting how environmental obstructions and reflections can significantly corrupt the received signal.

\section{Methodology}
\subsection{GNSS/Multisensor Fusion}
To develop and evaluate these methods, we utilize a novel multisensor state estimator based on continuous-time factor graph optimization (FGO) \cite{gnssfgo}, which serves as a flexible testbed for, e.g., deriving GNSS measurement residuals used in training and assessing the proposed uncertainty models. 

\subsection{Error Mitigation using M-Estimators}
In robust statistics, M-estimators offer a fixed-form, easily implemented approach that integrates prior information more simply than L- and R-estimates, while enabling simultaneous outlier detection during estimation \cite{de2021review}. 

Recalling the nonlinear least-square optimization formulated within a \emph{maximum-a-posteriori} (MAP) problem, where the cost function takes the form
\begin{align}
    \label{eq: l2_loss}
    J(\myFrameVec{e}{}{}) = \normsq{\myFrameVec{e}{}{}}{} \triangleq \normsq{\myFrameVec{z}{}{}-\myFrameVec{h}{}{}(\myFrameVec{x}{}{})}{\myFrameVec{\Sigma}{}{}}.
\end{align}
Under the Gaussian noise, the cost function in \eqref{eq: l2_loss} becomes quadratic, commonly referred to as the $L2$ (non-robust) loss. 

In this context, M-estimators aim to reshape the quadratic loss by introducing a symmetric, positive-definite kernel function $\rho(\cdot)$ with a unique minimum at zero. This modifies the original MAP problem as follows
\begin{align}
\label{eq: map_nls_m-est}
    \myFrameVecHat{X}{}{\rm MAP} = \arg\!\min_{\myFrameVec{X}{}{}} \sum_{k=0}^K \rho(\myFrameVec{e}{k}{}).
\end{align}
The kernel function $\rho(\cdot)$ in \eqref{eq: map_nls_m-est} is typically designed to reduce the influence of a large residual $\myFrameVec{e}{}{}$, which is often caused by outliers. Numerous kernel functions\footnote{Due to space limitations, we do not provide detailed technical explanations and instead refer readers to \cite{ZHANG199759, de2021review}.} are available in M-estimation, typically classified by their influence functions into three categories: \emph{robust-monotonic}, \emph{robust-soft-redescending}, and \emph{robust-hard-redescending}. 

Two key concepts, \emph{robustness} and \emph{relative efficiency}, are used to evaluate M-estimators, with a trade-off noted by \cite{Albuquerque}: \say{The more robust an estimator is, the less efficient it is}. Consequently, relative efficiency serves as a fundamental reference for tuning M-estimators.

For an in-depth study of handling GNSS outliers using M-estimators, we select six representative M-estimators, each parameterized at four efficiency levels, as summarized in Table \ref{tab: tuning_parameter}.

\begin{table}[!h]
\caption{Parameter $c$ of six M-estimators at four efficiencies.}
\label{tab: tuning_parameter}
    \centering
\resizebox{0.48\textwidth}{!}{\begin{tabular} {c|c|c|c|c||c}
     \hline\hline
    \diagbox{\textbf{Kernel}}{\textbf{Efficiency}}  &$\rm95\%$ &  $\rm90\%$ & $\rm85\%$ & $\rm80\%$ & Kernel Category \\
   \hline
    Fair & $1.3998$ & $0.6351$ & $0.3333$ & $0.1760$ & monotonic\\
    \hline
    Cauchy & $2.3849$ & $1.7249$ & $1.3737$ & $1.1385$ & soft-redescending\\
    \hline
     \begin{tabular}[c]{@{}c@{}} Geman-McClure \\ (GM)\end{tabular} & $3.8557$ & $2.8937$ & $2.5731$ & $2.2926$ & soft-redescending\\
    \hline
    Welsch & $2.9846$ & $2.3831$ & $2.0595$ & $1.8383$ & soft-redescending\\
    \hline
    Huber  & $1.3450$ & $0.9818$ & $0.7317$ & $0.5294$ & hard-redescending\\
    \hline
    Tukey & $4.6851$ & $3.8827$ & $3.4437$ & $3.1369$ & hard-redescending\\
    \hline\hline
\end{tabular}}
\end{table}

\subsection{Machine Learning for Outlier Prediction}\label{sec: nlos}
To address the limitations of current learning-based approaches for faulty GNSS prediction (see Section \ref{sec: literature_learning}), we take into account the noise characteristics of GNSS observations in urban environments and propose a transformer-enhanced Long Short-Term Memory (TE-LSTM) network in our previous work \cite{te_lstm}. 

We study the proposed network alongside a baseline LSTM network \cite{lstm_base} and two classical machine learning models: support vector machine (SVM) and extreme gradient boosting (XGBoost), aiming to provide a comprehensive analysis comparing classical models with deep networks, as well as evaluating performance on in-distribution and out-of-distribution data.

We investigate two learning tasks: NLOS classification and pseudorange error prediction as a regression task, using the input feature set
\begin{align}
    \mathbfcal{F} = \left\{\mathrm{El}_{t}^k,~\mathrm{Az}_t^k,~\mathrm{C/N_0}_t^k,~\sigma_{\mathrm{LS},t}^k,~\mathrm{RSS}^k \right\},
\end{align}
where the feature $\mathrm{El}$, $\mathrm{Az}$, and $\mathrm{C/N_0}$ represent the elevation and azimuth angle, and carrier-to-noise ratio of the $k$-th satellite at timestamp $t$, respectively.  The variable $\sigma_{\mathrm{LS}}$ denotes the pseudorange residual computed via an iterative least-squares method during pre-processing, which is further used to derive the root-sum-square (RSS) of pseudorange residuals within a specific time window.

\begin{figure*}[!t]
    \centering
    \includegraphics[width=1\textwidth]{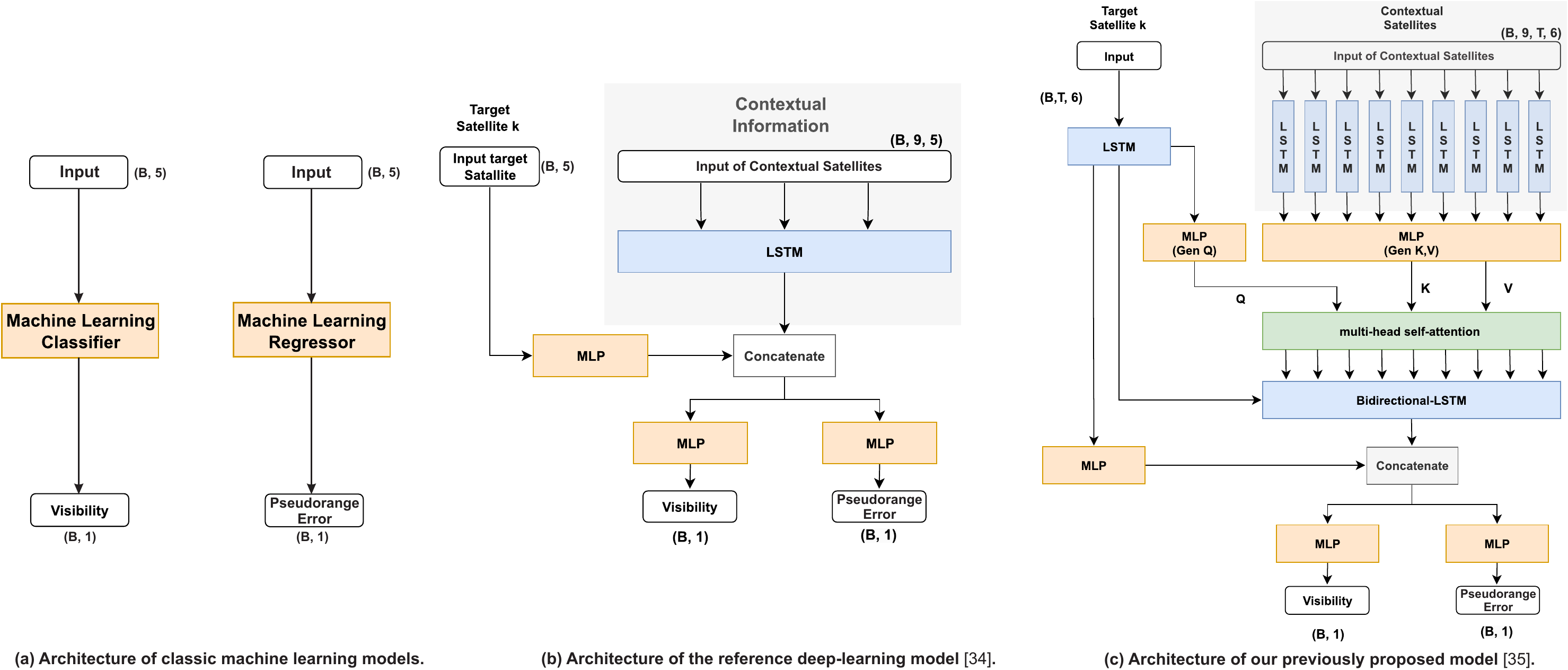}
    \caption{Learning models studied in this work.}
    \label{fig: learning_models}
\end{figure*}

Fig.\,\ref{fig: learning_models} illustrates the architectures of all learning-based models investigated in this work, where the variables $B$ and $T$ denote the batch size and time window length, respectively. The classical models are trained separately for the classification and regression tasks, whereas the deep learning models are trained jointly with corresponding loss terms for both tasks.

For a detailed explanation of all models, we refer the reader to \cite{te_lstm}.

\subsection{Bayesian Inference for Noise Distribution 
Approximation}\label{sec: gmm}
Following the previous work \cite{phd_Pfeifer}, we further explore the online learning paradigm by employing variational Bayesian inference to train a Gaussian Mixture Model (GMM) that approximates the pseudorange noise distribution\footnote{This work is currently in preparation for publication.}.

\begin{figure}[!t]
    \centering
    \includegraphics[width=0.4\textwidth]{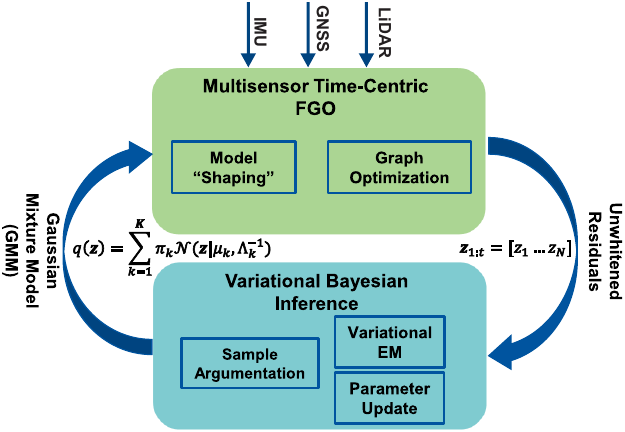}
    \caption{Concept of nested estimation with noise model update using variational Bayesian inference following \cite{phd_Pfeifer}.}
    \label{fig: gmm}
\end{figure}

Fig.\,\ref{fig: gmm} illustrates the concept of the proposed algorithm within a nested estimation framework. We employ multi-sensor factor graph optimization (FGO) to obtain \say{near-optimal} state estimates, which are then used to compute the pseudorange residuals. In a parallel stage, these residuals are sampled over a fixed time window, and a variational expectation-maximization (EM) algorithm is applied to approximate a GMM for each satellite \cite[Chapter~10]{Bishop06}. 

Due to limited space, we omit the detailed mathematical foundations of the inference algorithm and refer the reader to \cite{Bishop06} for a detailed description. Instead, we highlight two key challenges that extend beyond previous work \cite{phd_Pfeifer}, along with our proposed solutions.

The pseudorange residual $\varepsilon_{\rho,t}^{k}$ from satellite $k$ depends on the current state $\bm x_t$, while the GMM $\Theta_{t-1}^{k}$, estimated over $[t-1-\Delta t,~t-1]$, reflects a slightly outdated model. This temporal mismatch (distribution shift) is its main limitation.

In this context, we observed that two key issues arise from this. The first issue arises whenever the vehicle is in motion. As it moves relative to a mostly static environment, its position with respect to reflective surfaces changes, leading to varying residuals caused by the same structures. Although the Gaussian mixture model includes a component with a reasonable mean, it does not account for this relative positional change. Second, if the environment changes abruptly, the outdated GMM trained with past residuals may fail to capture new outliers, leading to poor state estimates. This creates a feedback loop: degraded estimates lead to inaccurate residuals, which further distort the GMM and worsen the next estimate.

To mitigate motion-induced distribution shifts, we propose Maximum Posterior Mean Augmentation (MPMA). 
The MPMA models a posterior over the mean-shift $\Delta \mu$ given a residual $\varepsilon$, using the dominant GMM component $\mathcal{N}(\mu_k, \Lambda_k^{-1})$:
\begin{align}
    p(\Delta \mu | \varepsilon) &\propto \mathcal{N}(\mu_k + \Delta \mu, \Lambda_k^{-1}) \cdot \mathcal{N}(0, \Tilde{\Sigma}) \label{eq3:14-15}
\end{align}
with $\Tilde{\Sigma} = \frac{1}{9}\Delta r_{j,t}^2$, ensuring 99\% of $\Delta \mu$ lies within probabilistic bounds.

To find the most probable mean shift $\left(\Delta \mu\right)^{\mathrm{MAP}}$, we take the negative logarithm of \eqref{eq3:14-15}, drop constant terms and obtain
\begin{equation}
    \left(\Delta \mu\right)^{\mathrm{MAP}} = \arg\min_{\Delta \mu} \mathcal{L}(\Delta \mu). \label{eq3:17}
\end{equation} 
By minimizing the convex loss
\begin{equation}
    \mathcal{L}(\Delta \mu) = (\varepsilon - \mu_k - \Delta \mu)^T \Lambda_k (\varepsilon - \mu_k - \Delta \mu) + \Delta \mu^T \Tilde{\Sigma}^{-1} \Delta \mu, \label{eq3:18}
\end{equation}
yielding the closed-form solution
\begin{equation}
    \left(\Delta \mu\right)^{\mathrm{MAP}} = \frac{\Lambda_k \Tilde{\Sigma} (\varepsilon - \mu_k)}{1 + \Lambda_k \Tilde{\Sigma}}. \label{eq3:19}
\end{equation}

To address the second issue, we extend the state estimator into a hybrid optimization framework inspired by \cite{dcsam}, where each pseudorange factor is modeled using a multi-hypothesis (MH) noise model:
\begin{equation}
    \eta(\varepsilon) = 
    \begin{cases}
        \sqrt{w(\Sigma^{-1/2}\varepsilon)}\, \Sigma^{-1/2}\varepsilon, & d = 0\\
        \Lambda_k^{1/2}(\varepsilon - \mu_k), & d = 1
    \end{cases},
    \label{eq3:23}
\end{equation}
with $\Sigma$ denoting the covariance of the Cauchy hypothesis, and $\mu_k$, $\Lambda_k \in \Theta$ representing the mean and precision of the dominant GMM mode. The discrete selector $d \in \{0,1\}$ determines the active noise model and is jointly optimized with continuous variables based on the following pseudo-probability
\begin{align}
\begin{split}
    &p(\varepsilon | d = 0) \\
    &=  \frac{1}{(2\pi)^{D/2}} |\Lambda_k|^{1/2} \left(1 - \exp \left\{ -\frac{1}{2} (\varepsilon - \mu_k^*)^T \Lambda_k (\varepsilon - \mu_k^*)\right\}  \right), \label{eq3:29}
\end{split}
\end{align}
where we reuse $\mu_k^{*}=\mu_k + \Delta\mu$ and $\Lambda_k$ from \eqref{eq3:19} and denote the residual's dimension by $D$, i.e. $\varepsilon \in \mathbb{R}^D$. 

\section{Data and Essential Experimental Results}
\subsection{Data Specification}
We conduct data from real-world measurement campaigns to train and test the investigated methods introduced above. Table \ref{tab: dataset3} summarizes the essential data characteristics. 

The data from Aachen (AC) captures a $\SI{17}{km}$ route through the city of Aachen, containing different driving scenarios: open-sky, semi-/dense-urban. The Hong Kong data (HK) comprises measurements from multiple satellite constellations, with up to 74 unique satellites observed across different timestamps. Data from Düsseldorf (DUS) with raw GPS and Galileo (GAL) measurements does not have labels and thus, are only used for testing.

\begin{table}[!h]
    \centering
    \caption{Data specification. The average, maximal, and minimum number of the received satellite are denoted with $n_{\mathrm{avg}}^{\mathrm{sat}}$, $n_{\mathrm{max}}^{\mathrm{sat}}$, and $n_{\mathrm{min}}^{\mathrm{sat}}$. $\sigma_{\mathrm{max}}^{\rho}$ denotes the maximal pseudorange error. $R_{}^{\mathrm{LOS}}$ and $R_{}^{\mathrm{NLOS}}$ represent the ratio of LOS and NLOS labels.}
   \resizebox{0.48\textwidth}{!}{
    \begin{tabular}{c|c|c|c|c|c|c|c}
    \hline\hline
     \textbf{Data} &  \begin{tabular}[c]{@{}c@{}}\textbf{Duration}\\ (\si{\second}) \end{tabular}  & $n_{\mathrm{avg}}^{\mathrm{sat}}$ & $n_{\mathrm{max}}^{\mathrm{sat}}$ & $n_{\mathrm{min}}^{\mathrm{sat}}$ & \begin{tabular}[c]{@{}c@{}}$\sigma_{\mathrm{max}}^{\rho}$ \\ (\si{\meter})\end{tabular} & \begin{tabular}[c]{@{}c@{}}$R_{}^{\mathrm{LOS}}$ \\ (\si{\percent})\end{tabular} &  \begin{tabular}[c]{@{}c@{}}$R_{}^{\mathrm{NLOS}}$ \\( \si{\percent})\end{tabular}\\
      \hline\hline
      HK   & 10573 & 16 & 25 & 4& 532.5 & 60.62 & 39.38\\
      \hline
      HK (GPS only)  & 10573 & 5 & 9 & 1& 276.08 & 60.47 & 39.53\\
      \hline
      AC (GPS only)   & 2366 & 6  & 8  & 0& 640.32 & 77.62 & 22.38\\
      \hline
      DUS (GPS+GAL) & 810  & 7  & 11 & 2 & 355.74 & - & - \\
      \hline\hline
    \end{tabular}
    }
    \label{tab: dataset3}
\end{table}

\subsection{Handling Outliers using M-Estimators}
To study the M-estimators, we divide the data from Aachen City into four sequences, as shown in Table \ref{tab: chap5_sequences}. 
\begin{table}[!h]
\caption{Sequence definition containing four scenarios in the dataset collected in Aachen.}
\label{tab: chap5_sequences}
\centering
\resizebox{0.48\textwidth}{!}{\begin{tabular} {c|c|c|c|c|c}
\hline\hline
\textbf{Seq.} & \textbf{Type} & \textbf{Location} & \textbf{Start} & \textbf{End} & \textbf{Specification} \\
\hline
1 & Open-Sky & Campus Melaten & $\SI{0}{s}$ & $\SI{150}{s}$ & no GNSS blockage\\
\hline
2 & Urban & Kaiser Platz & $\SI{750}{s}$ & $\SI{980}{s}$ & heavy GNSS blockage\\
\hline
3 & Sub-Urban & Europaplatz & $\SI{1200}{s}$ & $\SI{1400}{s}$ & slight GNSS blockage\\
\hline
4 & Sub-Urban & Eurogress & $\SI{1430}{s}$ & $\SI{1580}{s}$ & slight GNSS blockage \\
\hline
5 & Total route & - & $\SI{0}{s}$ & $\SI{2240}{s}$ & -\\
\hline\hline
\end{tabular}}
\end{table}

The error metrics of the mean $2$-D localization errors in our experiments are presented in Table \ref{tab: chap5_2D_error}. Recalling the definitions of robustness and efficiency, the error metrics reveal that the robustness of all M-estimators improves as efficiency decreases. Consequently, the positioning error decreases, provided that sufficient observations remain available to prevent the state estimator from diverging. 
\begin{table}[t]
\caption{Mean $2$-D error of all sequences is presented. Values marked in red indicate the worst accuracy, while those in green highlight the best results.}
\label{tab: chap5_2D_error}
\centering
\resizebox{0.48\textwidth}{!}{
\begin{tabular}{c|c||c||c||c||c||c}
\hline\hline
\diagbox{\textbf{Eff.}}{-}                   & \diagbox{\textbf{Ker.}}{\textbf{Seq.}} & \textbf{\begin{tabular}[c]{@{}c@{}}\#1 \\ (Open Sky)\end{tabular}} & \textbf{\begin{tabular}[c]{@{}c@{}}\#2 \\ (Urban)\end{tabular}} & \textbf{\begin{tabular}[c]{@{}c@{}}\#3 \\ (Light-Urban)\end{tabular}} & \textbf{\begin{tabular}[c]{@{}c@{}}\#4 \\ (Light-Urban)\end{tabular}} & \textbf{\begin{tabular}[c]{@{}c@{}}\#5\\ (Total Route)\end{tabular}} \\ \hline\hline
                                & Fair            & 0.35                                                               & {\color[HTML]{FE0000} \textbf{4.09}}                            & {\color[HTML]{FE0000} \textbf{1.74}}                                  & {\color[HTML]{FE0000} \textbf{3.08}}                                  & {\color[HTML]{CB0000} \textbf{2.51}}                                     \\ \cline{2-7} 
                                & Cauchy          & 0.35                                                               & 2.12                                                            & 1.05                                                                  & 1.60                                                                  & 1.38                                                                     \\ \cline{2-7} 
                                & GM              & {\color[HTML]{FE0000} \textbf{0.50}}                               & 0.67                                                            & 0.38                                                                  & 0.78                                                                  & {\color[HTML]{009901} \textbf{0.99}}                                     \\ \cline{2-7} 
                                & Welsch          & 0.47                                                               & {\color[HTML]{FE0000} \textbf{nan}}                             & {\color[HTML]{FE0000} \textbf{nan}}                                   & {\color[HTML]{FE0000} \textbf{nan}}                                   & {\color[HTML]{CB0000} \textbf{3.75}}                                     \\ \cline{2-7} 
                                & Huber           & 0.38                                                               & 3.26                                                            & 1.62                                                                  & 2.77                                                                  & 2.09                                                                     \\ \cline{2-7} 
\multirow{-6}{*}{\textbf{95\%}} & Tukey           & 0.38                                                               & 0.82                                                            & 0.95                                                                  & 0.69                                                                  & 1.11                                                                     \\ \hline\hline
                                & Fair            & 0.35                                                               & 2.97                                                            & 1.42                                                                  & 2.78                                                                  & 2.02                                                                     \\ \cline{2-7} 
                                & Cauchy          & {\color[HTML]{009901} \textbf{0.22}}                               & 0.76                                                            & {\color[HTML]{009901} \textbf{0.44}}                                  & 2.26                                                                  & 1.10                                                                     \\ \cline{2-7} 
                                & GM              & 0.30                                                               & 1.80                                                            & {\color[HTML]{009901} \textbf{0.41}}                                  & 1.80                                                                  & 1.20                                                                     \\ \cline{2-7} 
                                & Welsch          & {\color[HTML]{FE0000} \textbf{0.50}}                               & {\color[HTML]{009901} \textbf{0.50}}                            & {\color[HTML]{FE0000} \textbf{nan}}                                   & {\color[HTML]{FE0000} \textbf{nan}}                                   & {\color[HTML]{009901} \textbf{0.99}}                                     \\ \cline{2-7} 
                                & Huber           & 0.35                                                               & 2.78                                                            & 1.33                                                                  & 2.54                                                                  & 1.91                                                                     \\ \cline{2-7} 
\multirow{-6}{*}{\textbf{90\%}} & Tukey           & 0.38                                                               & 0.82                                                            & 0.95                                                                  & 0.69                                                                  & 1.11                                                                     \\ \hline\hline
                                & Fair            & 0.30                                                               & 1.90                                                            & 1.19                                                                  & 2.18                                                                  & 1.59                                                                     \\ \cline{2-7} 
                                & Cauchy          & 0.45                                                               & 1.13                                                            & 0.51                                                                  & 0.98                                                                  & 1.07                                                                     \\ \cline{2-7} 
                                & GM              & 0.31                                                               & 1.05                                                            & 1.25                                                                  & 0.64                                                                  & 1.15                                                                     \\ \cline{2-7} 
                                & Welsch          & 0.38                                                               & 0.82                                                            & 0.95                                                                  & 0.69                                                                  & 1.11                                                                     \\ \cline{2-7} 
                                & Huber           & 0.34                                                               & 2.30                                                            & 1.37                                                                  & 2.45                                                                  & 1.75                                                                     \\ \cline{2-7} 
\multirow{-6}{*}{\textbf{85\%}} & Tukey           & 0.38                                                               & {\color[HTML]{000000} 0.82}                                     & 0.94                                                                  & 0.69                                                                  & 1.11                                                                     \\ \hline\hline
                                & Fair            & 0.34                                                               & 1.72                                                            & 1.05                                                                  & 1.84                                                                  & 1.41                                                                     \\ \cline{2-7} 
                                & Cauchy          & 0.44                                                               & 0.73                                                            & 0.53                                                                  & 0.92                                                                  & {\color[HTML]{009901} \textbf{1.04}}                                     \\ \cline{2-7} 
                                & GM              & 0.36                                                               & {\color[HTML]{009901} \textbf{0.47}}                            & 0.69                                                                  & {\color[HTML]{009901} \textbf{0.46}}                                  & 1.09                                                                     \\ \cline{2-7} 
                                & Welsch          & {\color[HTML]{FE0000} \textbf{70.38}}                              & {\color[HTML]{FE0000} \textbf{nan}}                             & {\color[HTML]{FE0000} \textbf{nan}}                                   & {\color[HTML]{FE0000} \textbf{nan}}                                   & {\color[HTML]{CB0000} \textbf{70.38}}                                    \\ \cline{2-7} 
                                & Huber           & 0.31                                                               & 1.58                                                            & 1.25                                                                  & 2.08                                                                  & 1.49                                                                     \\ \cline{2-7} 
\multirow{-6}{*}{\textbf{80\%}} & Tukey           & 0.38                                                               & 0.82                                                            & 0.95                                                                  & 0.69                                                                  & 1.11                                                                     \\ \hline\hline
\end{tabular}
}
\end{table}

\subsubsection{Monotonous M-Estimator}
Monotonous M-estimators like Fair assign gradually decreasing influence to residuals. Fair rapidly increases influence for small errors and flattens beyond its tuning parameter, leading to over-penalization of mildly corrupted data. As a result, it shows consistently higher positioning errors. Additionally, its influence never reaches zero, reducing robustness compared to hard-redescending estimators.

\subsubsection{Hard-Redescending M-Estimator}
Hard-redescending estimators such as Tukey and Huber suppress outliers more aggressively. Tukey fully ignores large residuals, but its effectiveness heavily depends on the tuning parameter. Mis-tuning leads to valid data being discarded, limiting performance. Huber, though widely used, doesn't reject outliers outright and behaves more like Fair beyond the threshold. Neither shows consistent improvement over soft-redescending estimators in our tests.

\subsubsection{Soft-Redescending M-Estimator}
Soft-redescending estimators (Cauchy, GM, Welsch) offer a gradual reduction in influence for large residuals, balancing robustness and accuracy. Cauchy and GM perform best across all efficiencies (Table~\ref{tab: chap5_2D_error}), assigning reasonable weights even to smaller residuals potentially affected by noise. 
Welsch, however, showed instability, likely due to its exponential weighting, especially in urban scenarios. Overall, Cauchy and GM estimators provide the best trade-off between robustness and performance for range-based localization.

\subsubsection{\textbf{Discussions}}
Our findings highlight how to effectively apply M-estimators for handling measurement outliers, showing that aggressive down-weighting with low-efficiency estimators does not necessarily improve localization performance.

However, we do not propose a universal strategy for tuning hyperparameters, as generalizing M-estimation for long-term, large-scale localization remains challenging. 
Outlier characteristics and environmental variability create a difficult trade-off between robustness and efficiency in GNSS-based localization, where measurement corruption often coincides with signal loss. Even well-tuned estimators may fail to ensure reliable performance across diverse scenarios.

While prior work on generalized kernel functions and self-tuning estimators has improved generalization, consistent performance under dynamic GNSS conditions remains an open problem, especially when the noise characteristics of incoming measurements change abruptly—such as during multipath reception. These sudden shifts can disrupt the tuning process and degrade performance. 

This motivates a new open question: \say{\textbf{Can we predict measurement outliers using learning-based methods before integrating them into the estimator?}} Or more generally: \say{\textbf{Can we leverage learning to recover the information needed for tuning M-estimators?}}

\subsection{Handling Outliers using Learning-based Methods}
To evaluate the effectiveness and generalization of the learning-based methods in Section~\ref{sec: nlos}, we compare them with classical ML models (SVM, XGBoost) and deep learning baselines from \cite{lstm_base, te_lstm}, using the same dataset. Performance metrics are summarized in Table \ref{tab: metrics_ac_ac}\footnote{Due to limited space, we only present partial results and refer readers to \cite{te_lstm} for further details.}. 

\begin{table}[!t]
\centering
\caption{Performance Metrics using AC Dataset (reproduced from \cite{te_lstm}).}
\label{tab: metrics_ac_ac}
\resizebox{0.45\textwidth}{!}{
\begin{tabular}{c c c c c c}
               \hline\hline
               \textbf{Model} &&  \textbf{Precision} & \textbf{Recall} &\textbf{F1-Score} &\textbf{Accuracy}\\
                \cline{1-6}\hline   
                \multirow{2}{*}{SVM} &LOS &0.80&1.00&0.89 &\multirow{2}{*}{0.80}\\
                &NLOS &0.39&\textcolor{red}{0.01}&\textcolor{red}{0.01}\\
                \hline
                \multirow{2}{*}{SVM$^{*}$} &LOS &0.91&0.72&0.80 &\multirow{2}{*}{0.72}\\
                &NLOS &0.38&0.70&0.49\\
                \hline
                \multirow{2}{*}{XGboost} &LOS &0.86&0.88&0.87 &\multirow{2}{*}{0.79}\\
                &NLOS &0.47&0.42&0.45\\
                \hline
                \multirow{2}{*}{DL \cite{lstm_base}} &LOS &0.86&0.83&0.84 &\multirow{2}{*}{0.76}\\
                &NLOS &0.40&0.47&0.43\\    
                \hline
                \multirow{2}{*}{DL (ours, \cite{te_lstm})} &LOS &0.86&0.90&0.88 &\multirow{2}{*}{0.81}\\
                &NLOS &0.51&0.42&0.46\\
                \hline\hline
            
 \end{tabular}}
\end{table}

\begin{figure}[!t]
    \centering
    \includegraphics[width=0.5\textwidth]{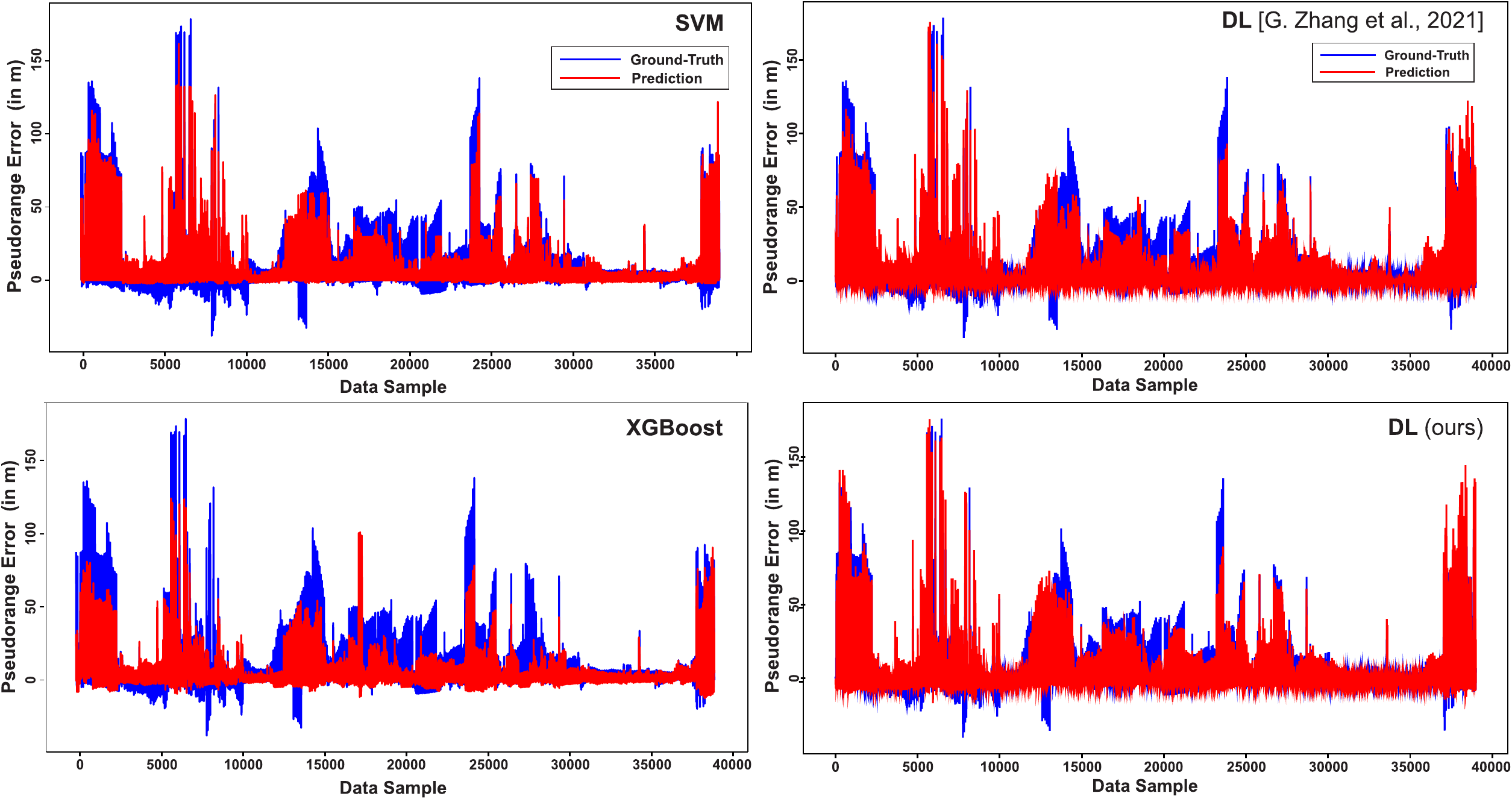}
    \caption{Results of pseudorange error prediction of all models using the HK data.}
    \label{fig: psr_error}
\end{figure}

\subsubsection{NLOS Detection}
The HK dataset, with a balanced LOS/NLOS ratio (Table~\ref{tab: dataset3}), served as an ideal benchmark. Classical models (SVM, XGBoost) performed well, slightly outperforming the deep learning baseline \cite{lstm_base} in precision and recall. Our reimplementation of the baseline yielded similar results, with slightly improved precision.

In contrast, the AC dataset posed a challenge due to its LOS-heavy imbalance. While all models maintained high LOS accuracy, NLOS classification dropped. SVM struggled most, with a recall of just $0.01$. Retraining on a balanced subset (SVM$^*$) significantly improved performance.

\subsubsection{Pseudorange Error Prediction}
In addition to NLOS classification, we evaluated all models on pseudorange error prediction, a more challenging task critical for robust GNSS positioning. 
We compared predicted and ground truth errors in Fig.\,\ref{fig: psr_error}, offering qualitative insights across error magnitudes. While all models captured general error trends, none reliably predicted large negative errors.

This limitation is concerning, as large negative errors can significantly degrade localization. Their rarity in the training data likely biases models toward small or positive errors, reducing generalization to critical, fault-prone cases.

\subsubsection{Feature Importance}
\begin{figure}[!t]
    \centering
    \includegraphics[width=0.45\textwidth]{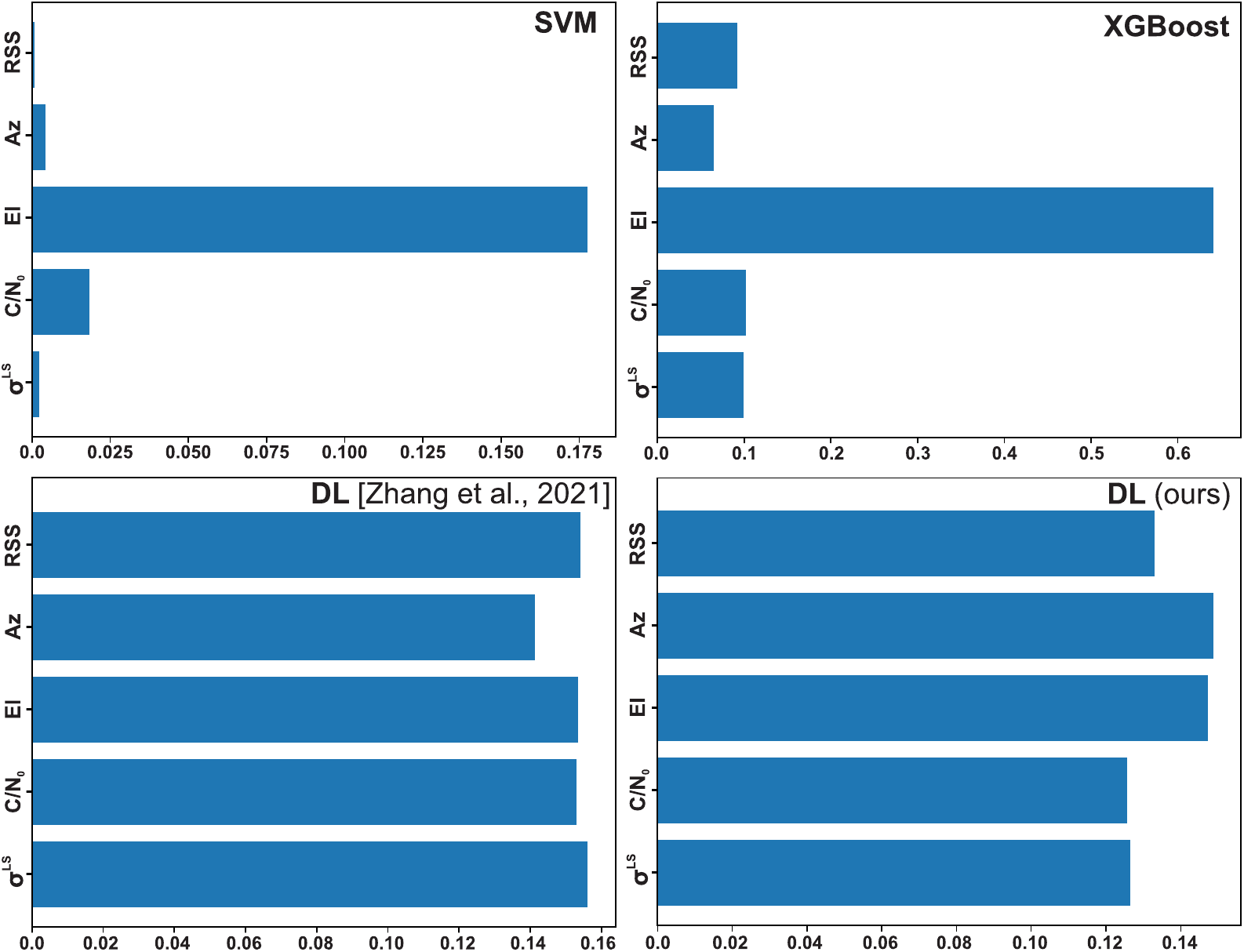}
    \caption{Feature Permutation Importance for NLOS classification (reproduced from \cite{te_lstm}).}
    \label{fig: feature_importance}
\end{figure}

To better understand model behavior, we analyzed permutation feature importance for all learning models (Fig.\,\ref{fig: feature_importance}). This highlights which features most influence predictions and how feature reliance affects generalization.

Classical models like SVM and XGBoost showed strong dependence on a few features, particularly elevation angle and $\mathrm{C/N}_0$. While informative, these features may overlook contextual cues crucial in urban environments. In contrast, deep learning models exhibited a more balanced reliance across features, indicating their ability to capture complex interactions and contextual patterns. 

\subsubsection{\textbf{Discussion}}

These results support the conclusion: \textbf{deep models generalize better}, though classical models excel on clean, balanced data. In addition, we think that learning-based methods are \textbf{not yet suited for back-end estimation but are effective for pre-processing}, aiding outlier detection and correction.

In addition, discriminative modeling with pre-trained models relies on a global model assumed to generalize across applications. This raises a key question: \say{\textbf{Does a truly global model for outlier characteristics exist?}}

Put differently: \say{\textbf{Do we need a global model at all?}} Or can robustness be enhanced through a \say{local} model that is updated in time and online? This leads us to the next method.

\subsection{Handling Outliers using Variational Bayesian Inference}
We implemented the proposed variational Bayesian inference to update the noise models in the pseudorange factors and evaluated its performance alongside Gaussian noise and the Cauchy estimator. Table \ref{tab4:7} summarizes the $2$-D error and FGO runtime statistics across the full dataset. In addition, Fig.\,\ref{fig: gmm} depicts the fitted mixture distribution alongside reference loss functions.

The Gaussian model lacks robust residual whitening, making it sensitive to outliers, which results in poor state estimates. Similarly, the standalone GMM model fails under severe outliers, preventing its use on the full dataset. Due to their limited robustness, both models are excluded from further analysis. In contrast, our MH-GMM with MPMA, which incorporates distribution-shift elimination, improves accuracy by $\SI{40}{\percent}$ compared to the M-estimator and by $\SI{11}{\percent}$ the naive MH-GMM.

\begin{table}[!t]
\caption{\label{tab4:7} Noise model statistics from a measurement campaign conducted along an $\SI{11}{km}$ route in Aachen.}
    \centering
    \vspace{-0.3cm}
\resizebox{0.45\textwidth}{!}{
\begin{tabular} {c|c|c|c|c}
    \hline\hline
    \multirow{2}{*}{\textbf{Noise Model}} & \multicolumn{2}{c|}{\textbf{2D Error (m)}} & \multicolumn{2}{c}{\textbf{Computation Time (ms)}}\\
    \cline{2-5}	
     & $\mathrm{mean}$ & $\mathrm{std}$ & $\mathrm{mean}$ & $\mathrm{std}$\\
    \hline
    Gaussian & $22.31$ & $28.49$ & $12.33$ & $5.16$  \\
    \hline
    M-estimator (Cauchy) & $0.90$ & $0.72$ & $\bm{11.93}$ & $4.88$  \\
    \hline
    GMM \cite{phd_Pfeifer} & \multicolumn{4}{c}{Failed}   \\
    \hline
    MH-GMM w. MPMA & $\bm{0.54}$ & $\bm{0.33}$ & $18.27$ & $6.17$  \\
    \hline
    MH-GMM w/o. MPMA & $0.61$ & $0.36$ & $17.87$ & $\bm{4.63}$  \\
    \hline\hline
\end{tabular}}
\end{table}

\begin{figure}[!t]
    \centering
    \includegraphics[width=0.48\textwidth]{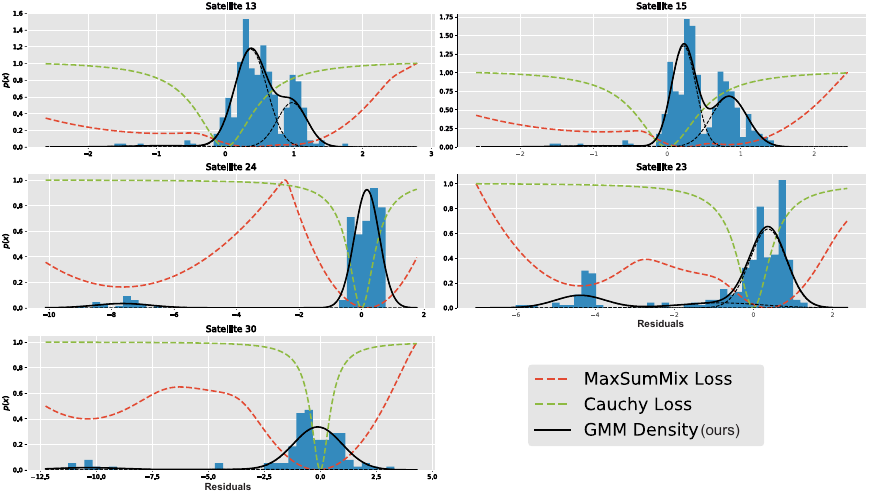}
    \caption{Demonstration of the online-approximated GMM distributions for different satellites while driving through a tree-rich area in Aachen.}
    \label{fig: gmm}
\end{figure}

\begin{figure}[!h]
    \centering
    \includegraphics[width=0.48\textwidth]{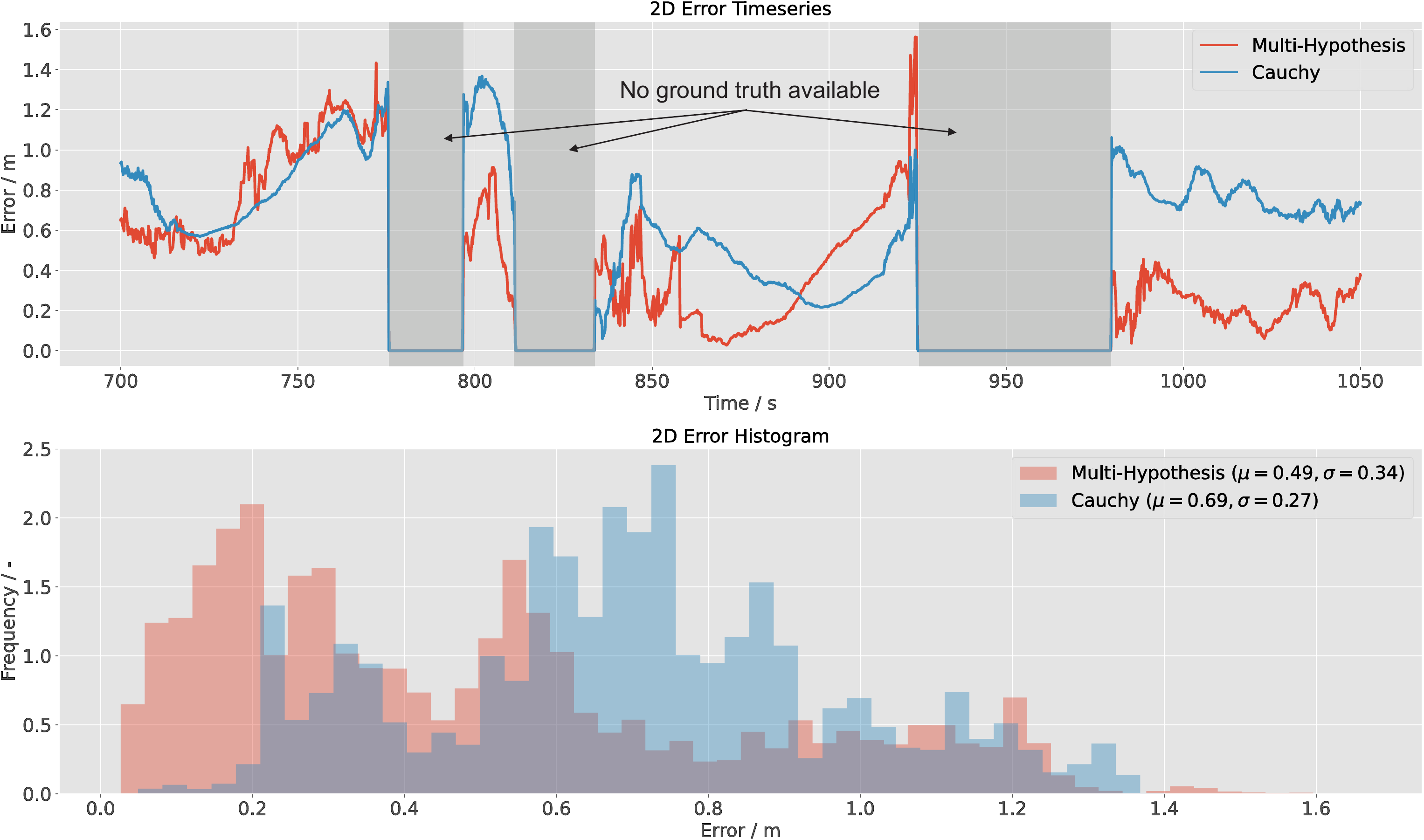}
    \caption{Plot of 2-D error and residual histogram in a highly urbanized area in Aachen (Kaiser Platz).}
    \label{fig: gmm_bushof}
\end{figure}

\textit{\textbf{Discussions}:} 
However, although the proposed inference method, updating only the noise models, can significantly improve the overall performance of vehicle localization, issues such as the smoothness of the estimated trajectory remain unresolved. Fig.\,\ref{fig: gmm_bushof} exemplifies this issue, showing that the estimator using MH-GMM exhibits high variations in 2-D errors due to unsmooth estimates. This may be caused by an inaccurate estimation of the switching variable $d$, which results in frequent transitions between different noise models.

Furthermore, we did not explicitly investigate \textbf{sample augmentations} in the current work, although this is crucial in the online learning setting. We also did not examine the \textbf{effect of the non-convexity of the approximated GMM} in solving the nonlinear least-squares problem, where alternative techniques such as Hessian approximation \cite{approx_hessian} could be explored.

\section{Conclusion}
Building on the results of this research, our objective is to explore various techniques for handling measurement outliers in range observations to enable robust vehicle localization in the field. Our methods, experimental evaluations, limitations, and open questions are summarized in this paper, providing a foundation for future work in enabling resilient vehicle localization in challenging field scenarios. 

Taken together, we emphasize the importance of outlier handling for achieving safe and reliable autonomy, highlighting the potential of learning-based methods and Bayesian inference as promising directions for future research.

\bibliographystyle{IEEETran.bst}
\bibliography{reference}
\addtolength{\textheight}{-12cm}   




\end{document}